# Inferring Strategies from Limited Reconnaissance in Real-time Strategy Games

**Jesse Hostetler** and **Ethan Dereszynski** and **Tom Dietterich** and **Alan Fern**
{hostetje, dereszet, tgd, afern}@eecs.orst.edu
Department of Electrical Engineering and Computer Science
Oregon State University
Corvallis, OR 97331

## Abstract

In typical real-time strategy (RTS) games, enemy units are visible only when they are within sight range of a friendly unit. Knowledge of an opponent's disposition is limited to what can be observed through scouting. Information is costly, since units dedicated to scouting are unavailable for other purposes, and the enemy will resist scouting attempts. It is important to infer as much as possible about the opponent's current and future strategy from the available observations. We present a dynamic Bayes net model of strategies in the RTS game *Starcraft* that combines a generative model of how strategies relate to observable quantities with a principled framework for incorporating evidence gained via scouting. We demonstrate the model's ability to infer unobserved aspects of the game from realistic observations.

## 1 INTRODUCTION

Real-time strategy (RTS) games, which are video game simulations of both the economic and military aspects of warfare, present a rich variety of reasoning challenges. Players must manage resources, plan at multiple time scales, coordinate heterogeneous armies composed of as many as 100 individual units, and respond to the actions of adversaries, all with incomplete information about the game state and under real-time constraints. We are particularly interested in the problem of decision-making with incomplete information. Specifically, how can an agent predict the intentions of its opponent from a limited number of observations?

This work addresses the problem of *unit count inference* during the early stages of the game. In the opening minutes, players make choices about what kinds of units to build and what technologies to pursue. These choices dictate the overall "feel" of the game: aggressive versus defensive, economy versus technology, etc. The players' plans are reflected in their opening *build orders*. Because resources are limited in the early game, players must commit to a particular path through the technology tree, culminating in a certain composition of the initial army. The primary concern early in the game is to choose an opening strategy that will gain an advantage against the opponent's opening.

This task is complicated by the limited information available to the players. Enemy units are visible only if a friendly unit is nearby. A player's knowledge of her opponent's strategy is thus limited to what has been observed through reconnaissance. It is impractical to observe all important actions of the enemy, because attempting to do so would divert many units from other tasks, and the enemy will attempt to destroy the scouts before they learn anything useful. Instead, players must gain whatever information they can with reasonable effort, and then evaluate their limited knowledge to infer the likely disposition of the enemy. An effective model of opening strategy must be able to make useful inferences from realistic evidence.

We present a dynamic Bayesian network model of opening strategy and apply it to the RTS game *Starcraft*. We combine a latent variable model of the true game state with an observation model that describes how the latent state generates the observed state. The true state, which we call the *battle space*, can be described by the true counts of each type of unit controlled by the enemy. The evolving counts of the various unit types are represented as Markov processes that are conditionally independent given the history of a hidden *strategy* process. The strategy determines which units are produced in a given time step, and the produced units accumulate to form the true counts. The observation model then determines the likelihood of observations given the true state and a measure of observation "effort." Although the hidden state space is potentially very large due to the number and car-dinality of the "count" variables, the structure of the model admits an efficient particle filtering algorithm for inference, taking advantage of the conditional independence of the unit counts.

## 1.1 REAL-TIME STRATEGY GAMES

We now briefly describe the mechanics of a typical RTS game. RTS games are simulations of total warfare. Players claim *bases*, which contain *resources*, by building *city centers*. They then use *workers* to harvest the resources, which they spend to construct *buildings* and *mobile units*. We will use the generic term *units* to refer both to buildings and to mobile units. *Production buildings* produce mobile units, including both workers and military units. In contrast, constructing *tech buildings* enables access to more powerful fighting units, and to further tech buildings. The game ends when one player controls no buildings, though players will typically resign a lost position much earlier.

Most RTS games contain a mechanism called the "fog of war". A metaphor for the limited intelligence available to military commanders, the fog of war makes enemy units invisible unless they are within visual range of a friendly unit. Players who lack accurate information about their opponents' activities are at risk of being surprised by attacks for which they are not prepared. Skilled players thus make scouting a high priority, and they are adept at inferring their opponents' intentions on the basis of that scouting. It is this inference process that we seek to model.

## 1.2 RELATED WORK

A complete description of a player's opening would specify how many units of each type existed at any moment. To our knowledge, no one has yet attempted to create a model of openings at this level of granularity. Existing work simplifies the task by adopting less-detailed descriptions of opening strategies.

A common simplification is to assume a small number of opening categories, such as "rush attack" or "strong economy." Weber and Mateas (2009) created a set of labels for *Starcraft* openings based on player-community vocabulary, and tried a variety of supervised classifiers for predicting the labels. Schadd et al. (2007) designed a two-level hierarchy of labels for the free RTS game *Spring*.

An alternative to categorical strategies is to model the dynamics of the game state itself. One can then represent an opening strategy as a path through the state space. Ponsen et al. (2007) used a finite state machine (FSM) model of building construction in *Wargus*. States represent which buildings have been built, and transitions are triggered by the construction of novel building types. Hsieh and Sun (2008) took a similar approach in *Starcraft* but used a stochastic FSM, with transition probabilities learned from gameplay data. Dereszynski et al. (2011) used a hidden Markov model (HMM). Transitions between hidden states occur at fixed time intervals, and the observations are Bernoulli random variables specifying the probability of building units of each type in the current state.

Except for Schadd et al. (2007), the work discussed so far assumes that the agent has access to information, such as the times at which particular units were constructed, that would be difficult to observe during actual gameplay. Another exception is Synnaeve and Bessière (2011), who used a directed graphical model to describe a joint distribution over strategy categories, game states, and observations in *Starcraft*. Their game states are Boolean vectors indicating whether each type of unit has been built, and game states are conditionally independent given strategy categories. Observations are incorporated by assigning 0 posterior probability to states that are inconsistent with observations.

The idea of an observation model (also known as a detection model) has been applied in many settings. Observation models are needed when we do not observe the true state of a system, but rather the results of some process parameterized by the latent true state. For example, in ecological modeling, we are often interested in the true number of individuals of a certain species present in the environment, but have access only to counts provided by field observers, who may not detect the species even if it is present (MacKenzie et al., 2002). The observation model describes how the hidden state determines the observed state, possibly as a function of covariates such as observer effort. One can then find the MAP estimate of the hidden state, accounting for the evidence. Our model is quite similar in spirit to the "occupancy-detection" models discussed by Hutchinson et al. (2011).

## 1.3 OUR CONTRIBUTIONS

This work improves upon existing models of RTS game opening strategy in several ways. First, we model the game state at the level of individual unit counts over time, enabling inference at the level of detail necessary for making gameplay decisions. Second, because individual unit counts are directly observable during gameplay, our model does not require evidence that is impossible to acquire in order to make inferences. Third, because our observation model includes predictors of scouting success, we can extract maximum value from limited observations, particularly from *failure* to observe a unit despite significant effort.

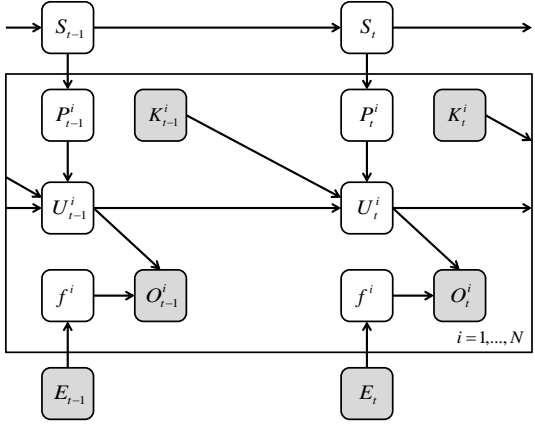

Figure 1: Two-slice representation of the reconnaissance model. We use plate notation to show that separate $P, K, U, f,$ and $O$ variables exist for each unit type $i$ (but note that the variables are not exchangeable). There is a direct link from $U_{t-1}^i$ to $U_t^i$ for each unit type. Shaded rectangles denote variables that we can directly observe.

## 2 THE MODEL

The overall model (Figure 1) is a discrete-time dynamic Bayes net (Dean and Kanazawa, 1988), combining two distinct components, the *state model* and the *observation model*, which we describe separately.

### 2.1 STATE MODEL

The state model describes the process that generates the true state of the battle space. Each game corresponds to a path through a discrete state space beginning at a designated start state. Let $S_t$ denote the state at time $t$. Time is discretized into 30-second intervals, and the player makes a state transition once every 30 seconds. We will say "epoch $k$" or "time $t = k$" to denote the time step begun by the $k$th transition. When the player visits state $s$, he or she decides how many units of each type to produce. Let $P_t^i$ be the number of units of type $i$ produced at time $t$. We model the decision to produce $k$ units of type $i$ as occurring in two steps. First, a decision is made about whether to produce any units of type $i$ at all. Then, if the first decision is to produce one or more units, a decision is made about how many units to produce. This is modeled by the zero-inflated Poisson distribution

$$P(P_t^i = k | S_t = s) = \begin{cases} 1 - \nu_s^i & k = 0 \\ \nu_s^i \cdot Pois(k-1; \lambda_s^i) & k > 0 \end{cases}$$

Here, $\nu_s^i$ is the probability of producing one or more units of type $i$ in state $s$, and hence, $1 - \nu_s^i$ is the probability of producing zero units. The number of units produced (beyond the first) is then determined by the Poisson rate parameter $\lambda_s^i$.

We model production as a two-step decision to make it easier to capture the decision not to produce any units of a particular type. This is important for modeling the production of *tech buildings*. In most cases, only one tech building of a given type is needed—it just serves as a prerequisite for producing higher-tech units. Decisions to produce tech buildings reflect strategic choices, so it is important to capture these 0/1 decisions. The alternative of just using a Poisson distribution does not handle this well, because there is no way to control the probability of producing 0 units without also changing the shape of the distribution. Similarly, modeling the unit production as Bernoulli random variables is not appropriate, because there do exist cases where two such buildings will be produced.

Let $\mathbf{P}_t = (P_t^i)_{i=1}^N$ denote the vector of unit production for all $N$ unit types at time $t$. In our model, the $P_t^i$ variables are independent conditioned on the current state $S_t$. Of course, there are important constraints among production rates of multiple unit types, since producing units of one type consumes resources that cannot be spent to produce other types. In principle, these constraints can be incorporated by using a sufficiently large number of states. However, that may conflict with the goal of capturing the player's overall strategy via the sequence of state transitions. Thus, the number of states must be selected appropriately to balance these objectives.

When scouting, we cannot directly observe the production of units; rather, a scout only observes (a subset of) the units that exist at a given time. To connect the individual unit production decisions to the scouting observations, we need to model the total number of units $U_t^i$ of type $i$ that exist at time $t$. This is equal to the number of units produced by time $t$ minus the number that have died. Each $U_t^i$ takes values in $\{0, 1, \ldots, U_{\max}\}$, where $U_{\max}$ is the maximum number of units that can exist. Because the production variable $P_t^i$ has infinite support, the assignment $U_t^i = U_{\max}$ represents the situation where there are $U_{\max}$ *or more* units of type $i$.

Unit deaths can occur in two ways. Most deaths involve our units killing their units, and these deaths, $\mathbf{K}_t = (K_t^i)_{i=1}^N$, are observed. Each unit of type $i$ also has a small probability $\ell^i$ of suffering an unobserved loss in each time step. For example, a building that was scouted while under construction may subsequently be canceled. Starting from initial unit counts $c^i$, the total count of unit type $i$ at time $t$ is defined recursively as

$$\begin{aligned} U_0^i &= c^i \\ U_t^i &\sim Binomial(U_{t-1}^i - K_{t-1}^i, 1 - \ell^i) + P_t^i \end{aligned}$$

When calculating $U_t^i$ in the case where $U_{t-1}^i = U_{\max}$,

we do not know the exact value represented by $U_{t-1}^i$, so we assume that it is *exactly* $U_{\max}$. Very rarely, this will lead us to underestimate $U_t^i$, but we set $U_{\max}$ to be large enough to avoid this problem in most cases.

Let $\mathbf{U}_t = (U_t^i)_{i=1}^N$ be the vector collecting the numbers of units of each type that exist at time $t$. We refer to $\mathbf{U}_t$ as the "battle space" at time $t$.

## 2.2 OBSERVATION MODEL

The *observation model* specifies the likelihood of an observation given a particular assignment to $\mathbf{U}_t$. We define an *observation* as the number of units of each type that we saw during a time interval, $\mathbf{O}_t = (O_t^i)_{i=1}^N$. We assume that we can distinguish between individual enemy units within an epoch, but not between epochs. That is, if we make two observations of a unit of a certain type during the same epoch, we know whether we have seen the same unit twice, or two different units. We can therefore model these observations as sampling without replacement from $\mathbf{U}_t$ within an epoch. For observations in different epochs, we assume that we cannot tell whether a unit observed in one epoch is the same as a unit observed in a previous epoch.

If we observe an enemy unit, we know that the unit exists. If, on the other hand, we do not observe a unit, there are two possible explanations. Either the unit does not exist, or it does exist but we did not look hard enough for it. Hence our observation model needs to incorporate some measure of scouting "effort". If we put little effort into scouting, failure to observe a unit does not tell us much about whether it exists. If we scout extensively and still do not see the unit, it probably does not exist.

We can exploit domain knowledge to come up with a measure of effort. In *Starcraft*, players construct most of their buildings in either their *main base* or their *natural expansion*. The main base is the area of the map where the player's starting units appear at the beginning of the game, while the natural expansion is the location where it is most "natural" to construct a second base. Because buildings must be defended, it is tactically advantageous to keep them close together. In the early game, the main base and the natural expansion are thus the most important areas to scout, since that is where the buildings will be located. A natural measure of scouting effort, then, is the proportion of these two areas that have been seen. We denote this proportion for slice $t$ as $E_t$.

We must now decide how our scouting effort influences the number of units we observe. Our initial approach was to treat each observable unit as an independent Bernoulli trial, with probability of success $E_t$. An observation would then be a vector of Binomial random variables, $O_t^i \sim Binomial(U_t^i, E_t)$.

The Binomial model assumes that the locations of units are distributed uniformly and independently in space. However, this assumption is wrong. Units tend to cluster together. For example, the primary task of "worker" units is to gather resources, which they do by traveling back and forth between the city center and the resource. Thus, almost all worker units will be found in the area between the city center and the resources. If we see one worker, it is probably because we have seen part of this area, and we would expect to have seen most of the other workers, too. There is thus more variance in the observations than the Binomial model would predict; the data are *overdispersed* with respect to the Binomial distribution.

We account for overdispersion by placing a Beta prior on the success probability parameter of the Binomial, forming a Beta-Binomial model (Haseman and Kupper, 1979):

$$O_t^i \sim BetaBinomial(U_t^i, \mu_t^i, \rho^i)$$
$$= \binom{U_t^i}{O_t^i} \frac{B(O_t^i + \alpha_t^i, U_t^i - O_t^i + \beta_t^i)}{B(\alpha_t^i, \beta_t^i)}$$

where $B(x, y)$ is the beta function, and

$$\alpha_t^i = \mu_t^i \frac{1 - \rho^i}{\rho^i}, \quad \beta_t^i = (1 - \mu_t^i) \frac{1 - \rho^i}{\rho^i}.$$

We adopt the $(\mu, \rho)$ parameterization of the Beta distribution, where $\mu \in [0, 1]$ is the mean of the Beta and $\rho \in [0, 1]$ is the dispersion parameter, which can be thought of as the correlation between individual successes. The Beta distribution is a conjugate prior of the Binomial. When $\rho \to 0$, the Beta-Binomial approaches the Binomial, while as $\rho$ increases, the density spreads out and eventually becomes bimodal. The bi-modality captures the "clumpiness" of the units: depending on whether the part of the area of interest we saw contains the units, the success probability is either high or low, but probably not in the middle.

For each unit type $i$, we learn a mapping $f^i(E_t)$ from the observation effort to the mean and dispersion of a Beta-Binomial distribution:

$$logit(\hat{\mu}_t^i) = a_0^i + a_1^i E_t$$
$$logit(\hat{\rho}^i) = b^i.$$

This is then plugged into the Beta-Binomial to compute the likelihood of observing $O_t^i$ given that $U_t^i$ units exist. We learn a different mapping for each unit type, to allow for differences in dispersion and ease of observability between unit types. The regression coefficients are assumed constant in time, but $\hat{\mu}_t^i$ varies in time due to its dependence on $E_t$.

## 2.3 TRAINING

All model variables except **S** are observed during training. Because all other variables are conditionally independent of **S** given **P** and **P** is observed, we can factor the full model into a latent variable model composed of **S** and **P**, and a fully observed model containing the rest of the parameters. We can learn the parameters of **S** and **P** separately from the rest, simplifying training.

The production process is a hidden Markov model (HMM) (Rabiner, 1990), where $S_t$ is the latent state and $P_t^1, ..., P_t^N$ are the emissions. The parameters of the HMM are the initial state probabilities $P(S_0) = Multinomial(\eta_1, ..., \eta_M)$, the state transition probabilities $P(S_t|S_{t-1} = s) = Multinomial(\pi_1^s, ..., \pi_M^s)$, and the inflated Poisson parameters $P(P_t^i = k > 0|S_t = s) = \nu_s^i \cdot Pois(k - 1; \lambda_s^i)$. We denote the parameter set of the HMM $\Phi = (\eta_1, ..., \eta_M, \pi_1^1, ..., \pi_M^M, \lambda_1^1, ..., \lambda_M^N, \nu_1^1, ..., \nu_M^N)$.

At training time, we observe $\mathbf{P}_t$. We can then estimate $\Phi$ in the usual way using the Expectation Maximization (EM) algorithm. We initialized the EM algorithm as follows: the $\eta$ and $\pi$ parameters are set to $1/M$, values of $\nu$ are drawn from a Uniform$(0, 1)$, and $\lambda$ parameters are drawn from a Uniform$(0, 10)$.

The "unobserved loss" probabilities $\ell^i$ are estimated as the number of unobserved losses of units of type $i$ divided by the number of unit-epochs (analogous to human-years) during which a unit of that type existed. For unit types that were present in at least 100 unit-epochs, $\ell^i$ was estimated using additive smoothing as $\hat{\ell}^i = \frac{d^i+1}{D^i+2}$ where $d^i$ is the number of unobserved losses of unit type $i$ and $D^i$ is the number of unit-epochs for type $i$. The smoothing ensures that all unit types have non-zero $\ell^i$ even if there were no unobserved losses in the training data. For types that were not present in at least 100 unit-epochs, the median estimate was used.

The functions $f^i$ giving the parameters of the distributions of $O_t^i$ are learned via logistic regression with a maximum likelihood objective using the R package aod (Lesnoff et al., 2010). We fit a $\hat{\mu}^i$ parameter for unit type $i$ only if a unit of that type was observed on at least 100 occasions in the dataset. We fit a $\hat{\rho}^i$ parameter only if the unit type met the condition for $\hat{\mu}^i$ and there were at least two of the unit type present (but not necessarily observed) on at least 100 occasions. The reason for the condition on $\hat{\rho}$ is that if it is rare for more than one instance of the unit to exist, then there is little dispersion in the data, and the estimate of $\hat{\rho}^i$ will be near 0. In this case, $\hat{\rho}^i$ would merely be modeling the tendency not to build more than one unit, which is properly the job of $\mathbf{P}_t$. For types that did not have enough data for $\hat{\mu}$ or $\hat{\rho}$, the median of the estimated regression coefficients are used.

## 2.4 INFERENCE

We denote the subset of latent variables for a slice $t$ as $X_t = \{S_t, P_t^1, ..., P_t^N, U_t^1, ..., U_t^N\}$, and the observed variables as $Y_t = \{E_t, K_t^1, ..., K_t^N, O_t^1, ..., O_t^N\}$. We use the lowercase $y_t$ and $x_t$ to denote instantiations of these variables (i.e, $y_t$ refers to the evidence at time $t$). Because each $U_t^i$ is conditioned on $S_t$, and $S_t$ and $U_t^i$ are Markovian, an exact filtering pass would require representing the forward message $\alpha_t = P(S_t, U_t^1, ..., U_t^N)$. This is intractable for even a modest number of types, since the size of the joint distribution is $MU_{\max}^N$. However, a key observation is that, given the history of the strategy state, $S_{0:t} = (S_0, ..., S_t)$, the model up to time $t$ decomposes into $N$ independent HMMs, each tracking the count of a single type. We leverage this structure by employing a Rao-Blackwellized particle filter (RBPF) for approximate inference (Doucet et al., 2000; Murphy, 2000).

In our application of RBPF, we draw particles of $S_{0:t}$, and compute $P(U_t^i|s_{0:t})$ analytically via standard HMM filtering. Following an importance sampling framework, particles are generated at each time step from a proposal distribution $Q(S_t)$. We use the state transition model for our proposal, $Q(S_t) = P(S_t|s_{t-1})$. While this choice ignores recent evidence at time $t$, it is computationally efficient to sample, and there are often periods of no evidence, anyway.

At $t = 0$ we draw $R$ particles from the initial state prior $s_0^1, ..., s_0^R \sim P(S_0)$. Each particle has an importance weight $w_t^r = \phi(s_t^r)/Q(s_t^r)$, where $\phi(s_t^r)$ is the probability of the particle's value $s_t^r$ given by the full model (up to normalization). At $t = 0$, $w_0^r = (P(y_0|s_r)P(s_r))/Q(s_r)$. For $t > 0$, each particle generates its next value of the state $s_t^i \sim Q(S_t) = P(S_t|s_{t-1}^i)$. We then update its weight using the ratio:

$$w_t^{r'} = \frac{P(Y_t = y_t|y_{0:t-1}, s_{0:t}^r)P(S_t = s_t^r|S_{t-1} = s_{t-1}^r)}{P(S_t = s_t^r|S_{t-1} = s_{t-1}^r)}.$$

The new weight for particle $r$ is then $w_t^r = w_{t-1}^r w_t^{r'}$.

Because our proposal distribution is identical to $P(S_t|S_{t-1})$ (canceling out the denominator), we are only interested in the likelihood term of the numerator, $P(Y_t = y_t|y_{0:t-1}, s_{0:t}^r)$, which factors as

$$\prod_{i=1}^N \Big[ P(U_t^i|U_{t-1}^i, P_t^i, K_{t-1}^i)P(U_{t-1}^{i,r}) \\ \cdot P(P_t^i|S_t = s_t^r)P(O_t^i|U_t^i, E_t^i) \Big].$$

$P(U_{t-1}^{i,r})$ is a forward-pass message that captures the posterior marginal distribution over the counts of unit type $i$ at time $t - 1$. After we weight a sample, we compute $P(U_t^{i,r}|s_{0:t}^r, y_t)$ to update its belief about the unit count and to pass to slice $t+1$. Each particle computes $N$ such messages, one for each unit type.

Particle filters often include a resampling step in which a new set of particles is sampled in proportion to the weights of the current set and given uniform weights. We do not include a resampling step because we expect that there will often be periods of no observations, particularly in the opening several epochs. Resampling would reduce the diversity of the particle population, with the result that if observations later come in that suggest *a priori* unlikely strategies, there may be no particles left that can represent them well.

## 3 EXPERIMENTS

We evaluated our model on its ability to infer the correct hidden unit counts, given the observations available to players during real games. We used two different metrics. For unit types that are usually present in numbers greater than 1, such as army units, we measured the model's ability to infer the correct *number* of units. For unit types that are either not built or built once, such as tech buildings, we measured accuracy in determining whether or not the unit is present. We also specifically tested the model's ability to infer the *absence* of units, a task that requires the observation model and the state model to work together.

Our experiments were conducted using gameplay data[1] from the RTS game *Starcraft*. We collected replays of 509 Protoss versus Terran games from the archives of the "Gosu Gamers" website.[2] *Starcraft* features three playable "races," Protoss, Terran, and Zerg, each with different units and abilities. We focused on a single match-up in this work. In our experiments, we take the perspective of the Terran player, and try to predict what the Protoss player is doing.

Data about unit counts and observations was extracted from the replays using the BWAPI library (BWAPI, 2012). We used the BWTA terrain analysis library (Perkins, 2010) to divide the maps into regions, and manually identified the regions corresponding to the Protoss player's main base and natural expansion.

The dataset was divided into 5 folds for cross-validation. To select $M$ (number of strategy states) in the production model, we compared the average likelihood $P(\mathbf{P}|\Phi)$ on the held-out data for $M = 20, 25, 30, 35$, and $40$. The likelihoods for $M = 25, 30$, and $35$ were statistically identical, and we selected $M = 30$ based on examination of the learned parameters and our knowledge of *Starcraft*. After selecting $M$, we estimated the observation model parameters for each of the 5 folds using their respective training sets.

[1]Our dataset is available at http://web.engr.oregonstate.edu/~tgd/rts/scouting/
[2]http://www.gosugamers.net/starcraft/replays/

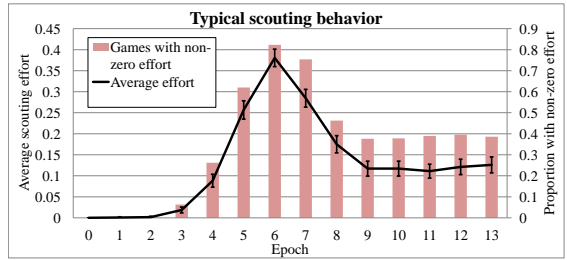

Figure 2: Average scouting effort, and proportion of games with non-zero effort, by epoch.

### 3.1 BASELINE METHOD

We implemented a simple baseline consisting of two sets of averages: the average number of units of each type at the end of each epoch for 1) all games, and 2) only games in which the unit was present in at least one epoch. At time $t$, if no units of type $i$ have been scouted so far, the baseline predicts the average for type $i$ at time $t$ across all games. If at least one unit of type $i$ has been scouted, the baseline predicts the average across games in which that unit type was present. The baseline should perform fairly well in the opening, since many openings are similar and thus unit counts often have low variance.

### 3.2 TYPICAL SCOUTING BEHAVIOR

Since our model uses observations obtained through scouting as its evidence, it is important to know what typical scouting behavior looks like. There is a pronounced peak in scouting effort from $t = 5$ to $t = 8$, after which effort falls back to a steady "background" level (Figure 2). The first significant scouting appears to begin at $t = 4$. Based on this pattern of scouting, we should expect our model to perform best during the peak scouting period, $t = 5, 6, 7, 8$.

### 3.3 MODEL ANALYSIS

The (logistic-transformed) regression coefficients of the observation model reveal varying levels of dispersion for different types of units. Typical values of $\hat{\rho}^i$ for mobile units were around 0.3, suggesting that mobile units often travel in groups. Values for buildings varied more widely. For Gateways (which produce military units), $\hat{\rho}^i$ was equal to 0.71, indicating that Gateways are very likely to be close to one another. At the other extreme, the value of $\hat{\rho}^i$ for the Nexus (the "city center" where resources are deposited by workers) was $10^{-8}$, since Nexi are never built near one another. The regression coefficients of the scouting effort were al-most all near 1.0, indicating that our scouting effort measure is a good predictor of scouting success. The two effort coefficients that were substantially less than 1 corresponded to units with the *Cloak* ability, which can only be seen by particular kinds of units. For one of these units, the Observer, the model learned that greater scouting effort *decreases* the probability of detection. This seems incorrect; we believe it occurs because in addition to being hard to detect, Observers usually are not fielded until after the period of peak scouting, so effort is negatively correlated with Observer presence.

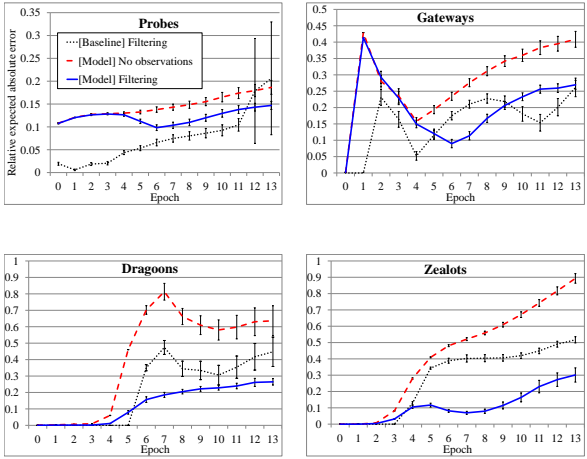

Figure 3: Relative expected absolute error for filtering predictions of common unit types, versus baseline. The error bars are 95% confidence intervals.

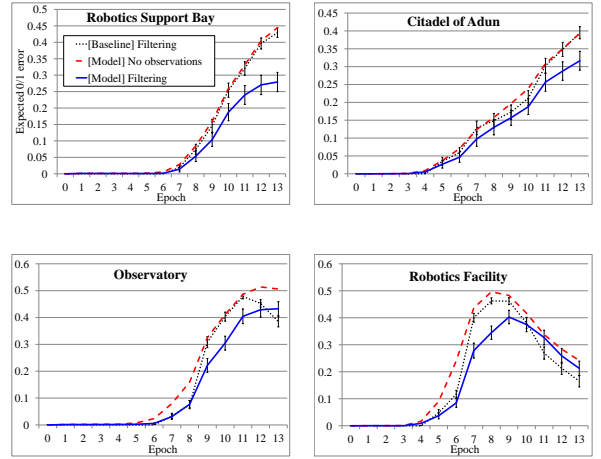

Figure 4: Expected 0/1 error for filtering predictions of tech buildings, versus baseline. The error bars are 95% confidence intervals. Error bars have been omitted for the model with no observations, for clarity.

### 3.4 INFERRING UNIT QUANTITIES

For common units that are built in large numbers, we would like to be able to infer the true counts. In *Starcraft*, the most common units in the early game are the Probe (the Protoss worker), the Dragoon and Zealot (basic military units), and the Gateway (a building that produces Dragoons and Zealots). Figure 3 compares our model's performance to that of the baseline for the filtering task over these four unit types. The dotted lines show the baseline method, the solid lines show our model's accuracy at filtering, and the dashed lines show our model's accuracy with no observations. The error measure is relative expected absolute error. That is, for a true count $u_t^i$ and the model's marginal distribution $U_t^i$, the error is given by

$$\varepsilon_t^i = E[|U_t^i - u_t^i|]/(u_t^i + 1).$$

We outperform the baseline from the onset of scouting for both Dragoons and Zealots. For Probes, we do worse for most of the opening, although both models have low error. We can attribute the good performance of the baseline to the low variance in Probe counts. Players almost always build Probes as quickly as possible in order to increase resource income, up to a "saturation" point that depends on the number of bases the player has secured. The large increase in the baseline's error at the end of the opening suggests that this is a point where some players have reached saturation while others have not. Our model does not experience a notable increase in error because it can represent multi-modal distributions. For Gateways, the baseline is notably better from $t = 10$ to $t = 12$. This is explained by the typical timing of Gateway construction. The first dip at $t = 4$ is a time when nearly everyone has built their first Gateway. Players will then build a second "batch" of Gateways, with the exact timing and number depending on their strategy. The second dip in baseline error around $t = 11$ is the end of the second period of Gateway production. Our model has trouble capturing this structure because it is stationary.

The shapes of the error curves for our model in the case of no observations are very similar to the baseline, although the baseline generally has better accuracy. Our model tracks the average-case behavior, but over-predicts production due to the Markov assumption on strategy states. For example, our model will give some probability of transitioning to a Dragoon-producing state earlier than a Dragoon could possibly have been produced.

### 3.5 INFERRING TECH BUILDINGS

While counts are important for units that are produced in numbers, for tech buildings we are primarily concerned with whether or not they exist. Thus, we compared our model to an appropriately modified baseline

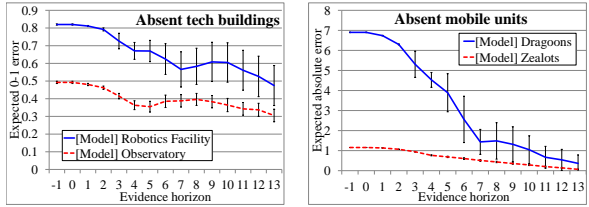

Figure 5: Error in predictions for $t = 13$ given evidence up to various horizons, on games where the target unit is *not* present. Error bars are 95% confidence intervals.

using expected 0/1 error for four tech buildings.

There are two qualitatively different shapes in the baseline error curves (Figure 4). For the Robotics Facility and the Observatory, baseline error peaked and then decreased, while for the Citadel of Adun ("Citadel") and Robotics Support Bay ("Support Bay"), it continued to increase for the entire opening. Our model with no observations tracks the baseline error closely, until the baseline error begins to drop.

We again seem to outperform the baseline from the onset of scouting, though the baseline sometimes makes a comeback at the end of the opening. As with the Gateway results in Section 3.4, the drop in error for the Robotics Facility at the end of the opening is explained by typical strategy. The great majority of openings involve building a single Robotics Facility before epoch 13. In constrast to the Gateway results, our model was able to capture this temporal structure for the Robotics Facility, perhaps because a Robotics Facility is built only once, whereas multiple Gateways are built over a period of time. On the other hand, a Support Bay will only be built in some openings, and will be skipped entirely in others. We would expect error to stabilize after the time period during which the Support Bay would be constructed passes, and we do see some evidence of this for our model with observations. It is also worth noting that the probability of scouting a Robotics Facility given that it exists is 0.37, while the probability of scouting a Support Bay is only 0.18.

### 3.6 INFERRING ABSENCE OF UNITS

Because we model the probability of observation success as a function of effort put into observing, we can make maximum use of *negative* observations to infer that units are *not* present. In Figure 5, we show our model's error for predictions of the *final* epoch, given evidence up to varying *horizons*, in games where the target unit was *not* present. For example, at horizon 6, the model is making predictions for $t = 13$ given evidence through $t = 6$. The true count is equal to 0, so all of the error comes from over-predicting. We see that as more negative observations arrive, the model revises its predictions downwards. The observation model uses the negative observations to infer that there are currently no units, and the rates of production given by the state model limit the number of units that could be built in the remaining time. The effect is strongest for units that are normally easy to scout. For example, Dragoons are present in all but 11 games, and they are easy to observe because they will be trying to attack the scout. Similarly, Robotics Facilities are both more common and more frequently scouted than Observatories. The baseline (not shown) would have a large, flat error on this task.

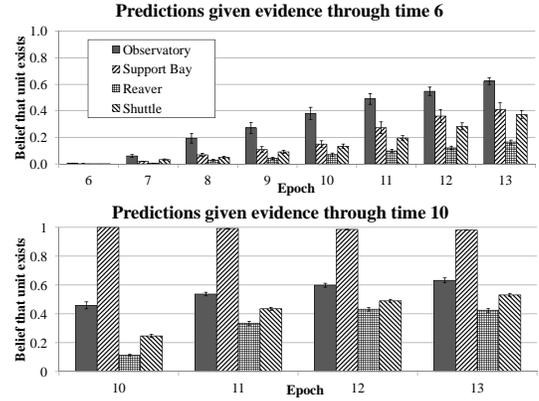

Figure 6: Belief that a unit of each of four different types will exist in future epochs in our Reaver drop case study game, given evidence up to $t = 6$ and $t = 10$. The Shuttle and Support Bay were actually built at $t = 9$, and the Reaver was built at $t = 11$. No Observatory was built. Error bars are 95% confidence intervals over 30 runs.

### 3.7 CASE STUDY

To demonstrate how our model would be applied in a game-playing agent, we examined a single game in detail. In this game, the Protoss player follows the *Reaver drop* strategy, which involves using a transport aircraft to carry a powerful unit called a Reaver behind enemy lines to attack the workers. This attack is potentially devastating, but easy to stop unless we are caught by surprise. We must note that one of the reasons that we chose this particular game is that the Terran player's scouting was particularly effective, giving our model a chance to make interesting inferences.

The key units in this strategy are the *Robotics Support Bay*, which is the tech building required for the Reaver; the *Shuttle*, which is the aircraft that transports the Reaver; and of course the Reaver itself. We can also contrast the Reaver drop opening to the "standard" opening, which involves building an *Observatory*.

This game features two distinct periods of scouting.

The Terran player scouts the Protoss at $t = 6$, leaves for a while, then returns to scout again at $t = 10$. Figure 6 shows, for each of the key units, the model's belief that at least one such unit will exist at each future time step, given evidence up to $t = 6$ and $t = 10$.

We first examine our model's predictions with evidence up to $t = 6$. The Terran player has just scouted the Robotics Facility, but the Support Bay and Reaver have not been built yet. Belief that an Observatory or a Support Bay exists begins to increase at $t = 8$. The Observatory is the "standard" continuation, so it receives more belief. Belief that a Shuttle exists increases in step with belief in the Support Bay because the Support Bay is strongly associated with the Reaver drop strategy, which always features a Shuttle. Belief that a Reaver exists increases more slowly than belief in the Support Bay because a Support Bay must be completed before a Reaver can be built. In the game, the Support Bay and Shuttle were actually built at $t = 9$, and the Reaver was built at $t = 11$.

We now examine how the model's predictions change when more evidence arrives. At $t = 10$, the Terran player scouts the Support Bay. This should unambiguously signal that a Reaver is going to be built, since there is no other reason to build a Support Bay. As expected, belief that a Reaver is coming increases considerably compared to the prediction with evidence through time 6. Belief that a Shuttle exists has also increased.

## 4 DISCUSSION

Our model generally performed well in comparison to the baseline for both count predictions and 0/1 predictions. The fact that our model is stationary while the baseline is non-stationary appears to account for most of the cases where the baseline was equal to or better than our model. Whereas the baseline can exploit the fact that certain patterns of production events are associated with particular time periods, in our model the Markov property of the hidden state erases information about how much time has elapsed. In the early game, our model begins predicting production events too early because it models the first several epochs, which all look the same, with a single state that has a high self-transition probability. Later in the game, uncertainty about the time means it has trouble capturing time-dependent "pauses" in production. On the other hand, our model can in principle be extended to full-length games, while our baseline method is too coarse to be useful much beyond the opening.

Another weakness of our model is that it incorporates no explicit prior knowledge about configurations of unit counts. For example, there will almost never be two Observatories, but our model can only account for this by designing the state transition matrix to visit the Observatory-producing state only once. Accuracy could be improved by making production of a unit at time $t$ ($P_t^i$) dependent on the count of that unit at time $t-1$ ($U_{t-1}^i$), with a corresponding increase in the complexity of the latent variable portion of the model. The situation worsens if we want to incorporate prerequisite relationships, as these cause the counting processes for *different* unit types to become coupled, destroying the conditional independences that we leveraged for efficient inference.

## 5 FUTURE WORK

A significant future challenge is to devise a model that can perform well in a full-length game. The space of possible strategies expands greatly as the game runs longer and the actions of the opponent begin to influence one's own decisions. Naively extending the current model by adding states is likely to prove intractable. One possibility is to try to exploit hierarchical structure in strategies to reduce the strategy space. We suspect that strategic decisions take place on multiple time scales—broad objectives at the top level, and smaller steps necessary to achieve them at the bottom. A model of a full game will need to incorporate a model of resource flow in order to reason effectively about production rates. It will also need to account for changes in the opponent's strategy in response to our actions.

We are also interested in applying opponent models to optimize scouting policies. As we saw in Figure 2, human players have converged on at least one time period in which the tradeoff between probability of scouting success and expected information gain is at an optimum. An agent could use our model to determine *when* important information is likely to be available. Modeling the probability that a scouting action will succeed in acquiring information, assuming that the information is there, is a challenging problem in itself.

### Acknowledgements

This work was made possible by the efforts of Mark Udarbe and Thao-Trang Hoang in assembling and labeling our dataset.

This work was partly funded by ARO grant W911NF-08-1-0242. The views and conclusions contained in this document are those of the authors and do not necessarily represent the official policies of ARO or the United States Government. Jesse Hostetler is partly supported by a scholarship from the ARCS foundation of Portland, OR.